# Evolving Chart Pattern Sensitive Neural Network Based Forex Trading Agents

Gene I. Sher
CorticalComputer@gmail.com
University of Central Florida

**ABSTRACT**
Though machine learning has been applied to the foreign exchange market for algorithmic trading for quiet some time now, and neural networks(NN) have been shown to yield positive results, in most modern approaches the NN systems are optimized through traditional methods like the backpropagation algorithm for example, and their input signals are price lists, and lists composed of other technical indicator elements. The aim of this paper is twofold: the presentation and testing of the application of topology and weight evolving artificial neural network (TWEANN) systems to automated currency trading, and to demonstrate the performance when using Forex chart images as input to geometrical regularity aware indirectly encoded neural network systems, enabling them to use the patterns & trends within, when trading. This paper presents the benchmark results of NN based automated currency trading systems evolved using TWEANNs, and compares the performance and generalization capabilities of these direct encoded NNs which use the standard sliding-window based price vector inputs, and the indirect (substrate) encoded NNs which use charts as input. The TWEANN algorithm I will use in this paper to evolve these currency trading agents is the memetic algorithm based TWEANN system called Deus Ex Neural Network (DXNN) platform.

**Categories and Subject Descriptors**
I.2.6 [**Computing Methodologies**]: Artificial Intelligence – Learning, Connectionism and neural nets.

**General Terms**
Algorithms

1. INTRODUCTION

Foreign exchange (also known as Forex, or FX) is a global and decentralized financial market for currency trading. It is the largest financial market, with a daily turnover of 4 trillion US dollars. The spot market, specializing in the immediate exchange of currencies, comprises almost 40% of all FX transactions, 1.5 trillion dollars daily. Because the foreign exchange market is open 24 hours a day, closing only for the weekend, and because of the enormous daily volume, there are no sudden interday price changes, and there are no lags in the market, unlike in the stock market. This paper presents the first of its kind, an introduction, discussion, analysis, and application/benchmarking (to the author's knowledge) of a topology and weight evolving neural network (TWEANN) algorithm for the evolution of geometry-pattern sensitive, substrate encoded trading agents that use the actual closing price charts as input. In this paper I will compare the Price Chart Input (PCI) using neural network (NN) based traders using substrate encoding, to the standard, direct encoded NN based trading agents which use Price List Input (PLI), in which the time series of closing prices is encoded as a list of said prices. Finally, all of these NN based trading agents will be evolved using the memetic algorithm based TWEANN system called Deus Ex Neural Network (DXNN)[1,2]. *It must be noted that it is not the goal of this paper to compare one TWEANN using PCI to another TWEANN using PCI. The use of DXNN system for this experiment and this paper was due to the ease with which it was possible to apply it to the given problem. The only goal of this paper is to demonstrate the utility of this new method, the use of candle-stick style chart as direct input to the geometry sensitive NN system evolved using a TWEANN system.*

The use of TWEANNs in the financial market has thus far been very seldom, and to this author's knowledge, the three most general of these state of the art neuroevolutionary algorithms (DXNN, NEAT[3], HyperNEAT[6], EANT[4], EANT2[5]) have not yet been thoroughly tested, benchmarked, and applied within this field.

In this paper I will not use the evolved NN based agents to predict currency pair prices, but instead evolve autonomously trading NN based agents. Neural networks have shown time and time again [7,8,9,10,11,12,13,14] that due to their highly robust nature, and universal function approximation qualities, that they fit well in the application to financial market analysis. In published literature though[15,17,18,19,20], the most commonly used neural network learning algorithm is backpropagation. This algorithm, being a local optimization algorithm, can and does at times get stuck in local optima. Furthermore, it is usually necessary for the researcher to set up the NN topology beforehand, and since the knowledge of what type of NN topology works best for which dataset and market is very difficult, or even impossible to deduce, one usually has to randomly create NN topologies and then try them out before settling down on some particular system. TWEANN systems are relatively new, and they have not yet been tested thoroughly in financial markets. But it is exactly these types of systems that can evolve not only synaptic weights, but also the NN topologies, and thus perform a global search, evolving the most optimal NN topology and synaptic weights at the same time. The use of such a TWEANN system in evolving NN based traders is exactly what will be explored in this paper.

Outline: In Section-2 I will briefly discuss Foreign Exchange market, the various market hypothesis, and the approaches that currency traders use to make their decisions. Section-3 will discuss how to use candlestick charts themselves as inputs to the NN based systems, and how to use substrate encoding[6] so that the NNs are sensitive to geometrical patterns present in these charts. In Section-4 I will introduce the memetic TWEANN system called DXNN, and discuss its various features and applicability to price prediction and automated trading, and how it will be used to evolve the direct and indirect encoded NN based currency trading agents. In Section-5 I will explain the benchmark setup for the testing of the PCI and PLI using NNs, and how I will gage fitness and generalization abilities of the evolved agents. In Section-6 I will present and discuss the benchmark results. Finally, in Section-7 I will conclude this paper with a summary, conclusions drawn, and proposed future work.

## 2. FOREIGN EXCHANGE MARKET

The foreign exchange market, or Forex, is a global, fully distributed, currency trading financial market. Unlike the stock market where a single buyer or seller with enough capital can dramatically change the price of the stock, the forex market is much too vast and distributed for any currency pair to be so easily affected. Furthermore, the fact that currencies can be traded non stop, 24 hours a day, 5 days a week, there are a lot fewer spaces in the data stream where news might be aggregating but no technical data is available. Because of these factors, there is a greater chance that the pricing data does indeed represent the incorporated news and fundamental factors, which might thus allow for prediction and trend finding through the use of machine learning approaches.

The question of predicting future market prices of a stock, or currency pairs as is the case in this paper, has been a controversial one, especially when using machine learning. There are two main market hypothesis which state that such predictions should be impossible. These two market hypothesis are the Efficient Market Hypothesis (EMH), and the Random Walk Theory (RWT).

The EMH states that the prices fully reflect all the available information, and that all new information is instantly absorbed into the price, thus it is impossible to make profits in the market since the prices already reflect the true price of the traded financial instrument. The RWT on the other hand states that historical data has no affect on pricing, and that the future price of a financial instrument is completely random, independent of the past, and thus it can not be predicted from it. Yet we know that profit is made by the financial institutions and independent traders in the existing markets, and that not every individual and institution participating in the trading of a financial instrument has all the available information immediately at his disposal when making those trades. Thus it can not be true that EMH and RWT fully apply in a non ideal system representing the real world markets. Therefore, with a smart enough system, some level of prediction above a mere coin toss, is possible.

There are two general approaches to market speculation, the technical and the fundamental. Technical analysis is based on the hypothesis that all reactions of the market to all the news, is contained within the price of the financial instrument. Thus past prices can be studied for trends, and used to make predictions of future prices due to the price data containing all the needed information about the market and the news that drives it. The fundamental analysis group on the other hand concentrates on news and events. The fundamental analyst peruses the news which cause the prices, he analyzes supply & demand, and other factors, with the general goal of acting on this information before others do, and before the news is incorporated into the price. In general of course, almost every trader uses a combination of both, with the best results being achieved when both of these analysis approaches are combined. Nevertheless, in this paper our NN systems will primarily concentrate only on the raw closing price data. Though in the future, the use of neuroevolution for news mining is a definite possibility, and research in this area is already in the works.

## 3. CREATING A GEOMETRICAL REGULARITY AWARE NEURAL NETWORK

Neural Networks have seen a lot of use and success in the financial market[17,18,19,20]. One of the main strengths of NN systems, which makes them so popular as market predictors, is that they are naturally non linear, and can learn non linear data correlation and mapping. Artificial neural networks are also data driven, can be on-line-trained, are adaptive and can be easily retrained when the markets shift, and finally, they deal well with data that has some errors; neural networks are robust.

When traders look at the financial data they do not usually look just at raw price lists, when a trader performs a time series analysis he instead looks at the chart patterns. This is especially the case when dealing with a trader prescribing to the technical analysis approach. The technical analyst uses the various technical indicators to look for patterns and emerging trends in these charts. There are many recurrent patterns within the charts, some of which have been given names, like for example the "head and shoulders" in which the time series has 3 hills, resembling head and shoulders. Other such patterns are the "cup and handle", the "double tops and bottoms", the "triangles"... Each of these geometrical patterns has a meaning to a trader, and is used by the trader to make predictions about the market. Whether these patterns really do have a meaning or not, is under debate. It is possible that the fact that so many traders do use these techniques, results in a self fulfilling prophecy, where a large number of the traders act similarly when encountering similar geometrical chart patterns, thus making that pattern and its consequence a reality, by all acting in a similar manner that is proscribed by the trend predicting rule of that pattern.

The standard neural networks used for price prediction, trend prediction, or automated trading, primarily use the sliding window approach, as shown in Fig-3.1, where the data is fed as a vector, a price list, to the NN. This vector, whether it holds only the historical price data, or also various other technical indicators, does not show these existing geometrical chart patterns which are used by the traders. If the NN does not have a direct access to the geometrical patterns used by human traders, it is at a disadvantage because it does not have all the information on which the other traders base their decisions.

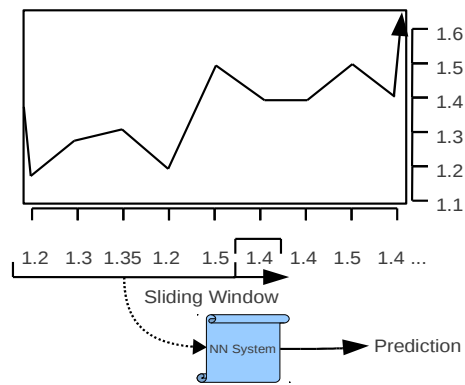

Fig-3.1 Sliding window (moving price list) based input

But how do we allow the NN to have access to this geometrical pattern data, and also be able to actually use it? We *can not* simply convert these charts to bitmaps and feed them to the NN, because the bitmap encoded chart will still be just a long vector, and the NN will not only have to deal with an input with high dimensionality (dependent on the resolution of the bitmap), but also there would really be no connection between this input vector and the actual geometrical properties of the chart that could be exploited.

A recently popularized indirect NN encoding approach that has been actively used in computer vision, and which has a natural property of taking geometrical properties of the input data into consideration, is the substrate (also known as hypercube) encoded NN system popularized by the HyperNEAT[6] implementation. In the substrate encoded NNs, the inputs and outputs are not fed directly to the neural network, but instead are fed into a substrate, in which the embedded neurodes (each possessing a coordinate) processes the signals and produces the outputs, as shown in Fig-3.2.

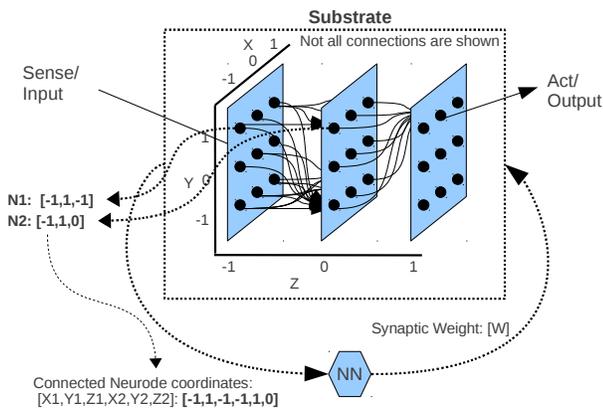

Fig-3.2 The substrate encoded neural network

As shown in the above figure, a substrate is a hypercube of neurodes, and though in the above figure the dimensionality is 3, it can be anything. Each neurode in one layer is connected in a feedforward fashion to the neurodes in the next layer, plane, cube, or hypercube, depending on the dimensionality of the entire substrate. All dimensions of the substrate have a cartesian coordinate, and the length of each side of the substrate is normalized such that the coordinates for each axis are between -1 and 1. The substrate is impregnated with neurodes, with the number of neurodes per dimension is set by the researcher, with each neurode having a coordinate based on its location in its substrate. With this setup, the weight that each neurode has for the presynaptic connection with other neurodes is then determined by the neural network to whom the substrate belongs. It is the neural network that calculates the synaptic weights for the connected neurodes based on the coordinates of those neurodes. The and pre and post-synaptic neurode coordinates are used as input to the NN. Because the NNs deal with coordinates, with the actual data input sent to the substrate, the substrate encoded NN system is aware of the geometric regularities within the input. And it is these geometric regularities that technical analysis tries to find and exploit.

With this type of indirect encoded neural network we can analyze the price charts directly, making use of the geometrical patterns, and trends within. Because each neurode in the substrate receives a connection from every neurode or input element in the preceding hyper-layer, the chart that is fed to the substrate must first be reconstructed to the resolution that still retains the important geometrical information, and yet is computationally viable as input. For example, if the sliding chart that is fed to the substrate is 1000x1000, which represents 1000 historical points (horizontal axis), with the resolution of the price data being (MaxPlotPrice – MinPlotPrice)/1000 (the vertical axis), then each neurode in the first hyperplane of the substrate will have 1000000 inputs. If the substrate has three dimensions, and we set it up such that the input signals are plane encoded and located at Z = -1, with a 10X10 neurons in the first neurode plane located at Z = 0, and 1X1 neurons in the third plane located at Z=1 (as shown in Fig-3.3), then each of the 100 neurons at Z = 0 receives 1000000 inputs, so each has 1000000 synaptic weights, and for this feedforward substrate to process a single input signal would require it 100*1000000 + 1*100 calculations, where the 1*100 calculations are performed by the neuron at Z = 1, which is the output neuron of the substrate. This means that there would be roughly 100000000 calculations per single input, per single processing price chart.

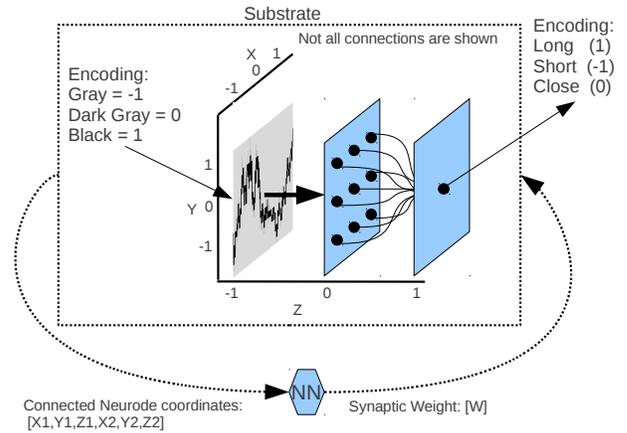

Fig-3.3 A layer-to-layer feedforward substrate processing a 2d chart input

Thus it is important to determine and test what resolution provides enough geometrical detail to allow for prediction to be made, yet not overwhelm the NN itself and the processing power available to the researcher. Once the length of the historical prices (horizontal axis on the price chart) and the resolution of the prices (vertical axis on the price chart) are agreed upon, the chart can then be generated for the sliding window of currency pair prices, producing the sliding chart. For example, Fig-3.4A shows a single frame of the chart whose horizontal and vertical resolution is 100 and 20 respectively, for the EUR/USD closing prices taken at 15 minute time-frame (pricing intervals). This means that the chart is able to capture 100 historical prices, from N to N-99, where N is the current price, and N-99 is the price (99*15)min ago. Thus, if for example this chart's highest price was $3.00 and the lowest price was $2.50, and we use a vertical resolution of 20, the minimum discernible price difference (the vertical of the pixel) is (3-2.5)/20 = $0.025. For comparison, Fig-3.4B shows a 10x10 chart resolution.

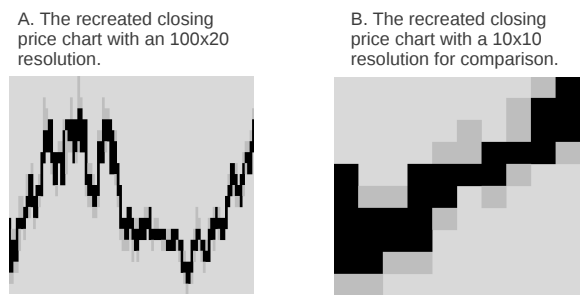

A. The recreated closing price chart with an 100x20 resolution.

B. The recreated closing price chart with a 10x10 resolution for comparison.

Fig-3.4 A. and B. show a 100x20 and 10x10 resolution based charts respectively, using the candle-stick charting style.

Similar to Fig-3.3, in Fig-3.4 the pixels of the background gray are given a value of -1, the dark gray have a value of 0, and the black a value of 1. These candlestick charts, though of low resolution, retain most of the geometrical regularities and high/low pricing info of the original plot, with the fidelity increasing with the recreated chart's resolution. It is this input plane that can be fed to the substrate.

## 4. Deus Ex Neural Network Platform

The topology and the synaptic weights of the neural networks need to be set and optimized for the tasks the NNs are applied to. One of the strongest approaches to the optimization of synaptic weights and their topologies is through the application of evolutionary algorithms. The systems that evolve both the topology and the synaptic weights of neural networks are called Topology and Weight Evolving Artificial Neural Networks (TWEANN). DXNN is a memetic algorithm based TWEANN system, and is the algorithm I will use to evolve the NN based currency trading agents in this paper. In this section we will briefly discuss DXNN's various features, and what makes it different from other TWEANNs.

### 4.1 The memetic approach to synaptic weight optimization

The standard genetic algorithm performs global and local search in a single phase. A memetic algorithm separates these two searches into separate stages. When it comes to neural networks, the global search is the optimization and evolution of the NN topology, while the local search is the optimization of the synaptic weights.

Based on the benchmarks, and ALife performance of DXNN[1], the memetic approach has shown to be highly efficient and agile. The primary benefit of separating the two search phases is due to the importance of finding the right synaptic weights for a particular topology before considering that topology to be unfit. In standard TWEANNs, a system might generate an optimal topology for the problem, but because during that one innovation of the new topology the synaptic weights make the topology ineffective, the new NN topology is discarded. Also, in most TWEANNs, the synaptic weight perturbations are applied indiscriminately to all neurons of the NN, thus if for example a NN is composed of 1 million neurons, and a new neuron is added, the synaptic weight mutations might be applied to any of the now existing 1000001 neurons... making the probability of optimizing the new and the right neuron and its synaptic weights, very low. The DXNN platform evolves a new NN topology, and then through the application of an augmented stochastic search with random restarts optimizes the recently added synaptic weights for that topology. Thus when the "tuning phase", as is the local search phase is called in the DXNN platform, has completed, the tuned NN has roughly the best set of synaptic weights for its particular topological architecture, and thus the fitness that is given to the NN is a more accurate representation of that NN's true potential. Furthermore, because the synaptic weight optimization through perturbation is not applied to all the neurons indiscriminately throughout the NN, but instead is concentrated on the newly mutated, augmented, or mutationally affected neurons, the tuning phase optimizes the new additions to the NN, making those new elements work with the existing, already evolved and proven architecture. Combined together, the DXNN's approach to neuroevolution tends to produce a more efficient and concise NN systems. The benchmarks in paper [1] demonstrated it to rapidly evolve neurocontrollers for agents in the ALife simulation, which gives hope that this neuroevolutionary system is also powerful enough to produce positive results in this application, and is the reason DXNN was chosen for this paper.

### 4.2 The DXNN Neuroevolutionary Process

The DXNN evolutionary algorithm performs the following steps:
1. Create a seed population of topologically minimalistic NN genotypes.
2. **Do:**
   1. Convert the genotypes to phenotypes.
   2. **Do for every NN (Apply parametric tuning):**
      1. Test the fitness of the NN system.
      2. Optimize the synaptic weights through the application of synaptic weight tuning (An augmented version of stochastic hill-climbing algorithm).
      
      **Until:** The fitness has failed to increased K number of times.
   3. Convert the NN system back to its genotype, with the tuned synaptic weights, and its fitness score.
3. After all the NNs have been given a fitness score, sort the NN agents in the population based on their fitness score, which is further weighted based on the NN size, such that smaller sized NNs are given priority.
4. Delete the bottom 50% of the population.
5. For each NN, calculate the total number of N offspring that it is alloted to produce, where N is proportional to the NN's fitness as compared to the average fitness of the population, and average NN size, where the smaller and more fit NNs are allowed to create more offspring.
6. Create the offspring by first cloning the fit parent, and then applying to the clone T number of mutation operators, where T is randomly chosen to be between 1 and sqrt(Parent_TotNeurons), with uniform probability. Larger NNs will produce offspring which have a chance to be produced through a larger number of applied mutation operators.
7. Compose the new population from the fit parents and their offspring.

**Until:** Termination condition is reached (Max number of evaluations, time, or goal fitness).

What in the DXNN is referred to as a "tuning phase" is the local search phase of the algorithm, which as noted is an augmented stochastic search algorithm. The topological mutation phase, by randomly choosing the number of mutation operators to use when producing offspring by applying the said mutation operators to the clones of the fit parents, allows for a high variability of topological mutants to be created, improving the diversity of the population. The DXNN system uses the following list of mutation operators:
1. Add new neuron.
2. Splice two neurons (choose 2 connected neurons, disconnect them, and then reconnect them through a newly created neuron. This also increases the depth of the NN).
3. Add an output connection to a randomly selected neuron, recurrent or feedforward.
4. Add an input connection to a randomly selected neuron.
5. Add a sensor
6. Add an actuator

Thus through mutation operators 5 and 6, the offspring might incorporate into itself new and still unused sensors and actuators, if those are available in the list of sensors and actuators for its specie/population. Indeed this particular part of the DXNN acts as a natural feature selection, and is especially useful for complex problems, in robotics, and alife simulations. In alife in particular, as was shown in [1], the organisms were able to evolve and integrate new sensors over time. This can also be used in robotics, letting evolution decide what sensors and actuators are most useful to the evolving individual. But more importantly, this can be used in evolving algorithmic trades, where Sensors can represent the different types of technical indicators.

Furthermore, DXNN evolves both, direct and indirect (substrate in this case) encoded NNs. Where its substrate encoded NNs further differ in their ability to evolve different types of coordinate based preprocessors, which is hoped to allow it to deal with a more varied number of geometrical features, and was another reason for choosing it in this experiment. A further elaboration on this is discussed next.

### 4.3 Direct and Indirect Encoding

The DXNN platform evolves both direct and indirect encoded NN systems. The direct encoded NN systems were discussed in the above sections, the indirect encoded NNs use substrate encoding. Since it is the substrate that accepts inputs from the environment and outputs signals to the outside world, and the NN is used to set the synaptic weights between the neurodes in the substrate, the system not only has a set of sensors and actuators as in the standard NN system, but also a set of "coordinate_preprocessors" and "coordinate_postprocessors", that are integrated into the NN during evolution in a similar manner that it integrates new sensors and actuators, only using the "add_coord_preprocessor" and "add_coord_postprocessor" mutation operators.

In the standard substrate encoded NN system, the NN is given an input that is a vector composed of the coordinates of the neurode for which the synaptic weight must be generated, and the coordinates of the presynaptic neurode which connects to it. In DXNN, there are many different types of coordinate_preprocessors and coordinate_postprocessors available. The coordinate_preprocessors calculate values from the coordinates passed before feeding the resulting vector signals to the NN. The coordinate_postprocessors post process the NN's output, adding plasticity and other modifications provided by the particular substrate_actuator incorporated into the system through evolution.

Wheres HyperNEAT which popularized substrate encoding, feeds the CPNN (a NN that uses tanh, and other types of activation functions) simply the coordinates of the presynaptic and postsynaptic neurodes, and in some variants the distance between the neurodes, the DXNN uses the following list of coordinate_preprocessors. All of these are available to the substrate encoded NNs in this paper, and can be integrated and used by the substrate encoded NN as it evolves:
1. Cartesian Coordinates
2. Cartesian Distance
3. Convert to Polar (if substrate is a plane)
4. Convert to Spherical (if substrate is a cube)
5. Centripetal distance of neurode
6. Distance between coordinates
7. Gaussian processed distance between coordinates

This set of substrate sensors further allow the evolved NN to extract geometrical patterns and information from inputs. Also, DXNN allows the evolved neurons to use the following list of activation functions: *[tanh,gaussian,sin,absolute,sgn,linear,log, sqrt]*, whether those neurons are used by standard direct encoded NNs, or NNs used to calculate weights for the substrate encoded systems. Which the creator of DXNN hopes will further improve the generality of the evolve NN systems, and has indeed shown benefit when evolving standard neurocontrollers for the double-pole balancing benchmark, in which the problems were solved faster by NNs that used sinusoidal activation functions.

### 5. The Experimental setup

The 2 goals of this paper is to test the applicability and effectiveness of TWEANN systems in the evolution of automated currency trading NN based agents, and the testing and comparison of the effectiveness/profitability and generalization properties of the Price List Input (PLI) based NNs, and the Price Chart Input (PCI) used by geometrical pattern aware substrate encoded NNs. The hypothesis is that the PCI NNs will have much more information (Actual geometrical properties of the chart, including the relative pricing) than the standard PLI based NNs, which do not have access to the geometrical properties of the charts.

For this benchmark I created a forex market simulator, where each interfacing NN will be given a $300 starting balance. Each agent produces an output, with the output being further converted to **– 1 if its less than -0.5**, **0 if between -0.5 and 0.5**, and **1 if greater than 0.5**. When interacting with the forex simulator, -1 means go short, 0 means close position (or do nothing if no position open), and 1 means go long (if currently shorting, then close the position, and then go long). The Forex simulator will simulate the market using 1000 real EURUSD currency pair closing prices, stretching from 2009-11-5-22:15 to 2009-11-20-10:15, with 15 min time frame (each closing price is 15 minutes from the other). The simulator uses a price spread of $0.00015.This 1000 point dataset is split into the training set, and into a testing/generalization set. The training set is the first 800 time steps, ranging from: 2009-11-5-22:15 to 2009:11-18-8:15, and the testing/generalization data set is the immediately following 200 time steps from 2009-11-18-8:15 to 2009-11-20-10:15. Finally, when opening a position, it is always done with $100, leveraged by x50 to $5000 (due to the use of a flat spread and buy/sell slots, the results can be scaled).

A single evaluation of a NN is counted if the NN based agent has went through all the 800 data points, or if its balance dips below $100. The fitness of the NN is its networth at the end its evaluation. Each evolutionary run lasts for 25000 evaluations, and each experiment is composed of 10 such evolutionary runs. In each experiment the population size was set to 10. Finally, in every experiment the NNs were allowed to use and integrate through evolution the following set of activation functions: *[tanh, gaussian, sin, absolute, sgn, linear, log, sqrt]*. The remainder of the parameters were set to the values recommended in [1].

In the experiments performed, the NNs used price sliding window vectors (for direct encoded NNs), and price charts (for recurrent substrate encoded NNs) as shown in Fig-5.1. The NNs were also connected to a sensor which fed them the vector signal: [Position,Entry,PercentageChange], where Position takes the value of either -1 (currently shorting), 0 (no position), or 1 (currently going long), Entry is the price at which the position was entered (or set to 0 if no position is held), and PercentageChange is the percentage change in the position since entry.

In this paper I present 13 benchmarks/experiments, each experiment is composed of 10 evolutionary runs from which its average/max/min are calculated. The experiments demonstrate and compare the performance of PCI based NNs and the PLI based NNs. Both these input type experiments were tested with different sensors of comparable dimensionality.

- **5 PLI experiments:**

Experiments 1-5 were performed using with PLI NNs. Each experiment differed in the resolution of the sliding window input the NNs used. Each NN started with the sliding window sensor, and the vector: [Position, Entry, PercentageChange]. The networks were allowed to evolve recurrent connections. These 5 experiments are:
1. [SlidingWindow5] 2. [SlidingWindow10] 4. [SlidingWindow20] 4. [SlidingWindow50] 5. [SlidingWindow100]

- **8 PCI experiments:**

Experiments 6-13 were preformed using the PCI NNs. In these experiments each PCI based NN used a 4 dimensional substrate. The input hyperlayer to the substrate was composed of the price chart, the vector: [CurrentPosition, EntryPrice, PercentageChange], and the substrate's own output, making the substrate **Jordan Recurrent**. The substrate architecture of these CPI NN based agents is shown in Fig-5.1. The reason for using Jordan Recurrent substrates is due to the fact that standard feedforward substrates which do not have recurrent connections, though achieving high fitness during training, did not generalize almost at all during my preliminary experimentation, with the highest achieved balance during generalization testing phases being $303 (a $3 profit), but usually dipping below $250 (a $50 loss) during most evolutionary runs. Thus for the PCI based NNs, I created a 4 dimensional substrate (the 4$^{th}$ dimension was called K) with an input hyperplane composed of the noted 3 planes and located at K = -1, all of which connected to the 5X5 hidden plane positioned at K = 0, which then further connected to the 1X1 output plane (a single neurode) located at K = 1, which output the short/hold/long signal and which was also fed back to the substrate's input hyperplane. Each of the 10 experiments used price chart inputs of differing resolutions:
1. [ChartPlane5X10], 2. [ChartPlane5X20] 3. [ChartPlane10X10]
4. [ChartPlane10X20] 5. [ChartPlane20X10] 6. [ChartPlane20X20]
7. [ChartPlane50X10] 8. [ChartPlane50X20].

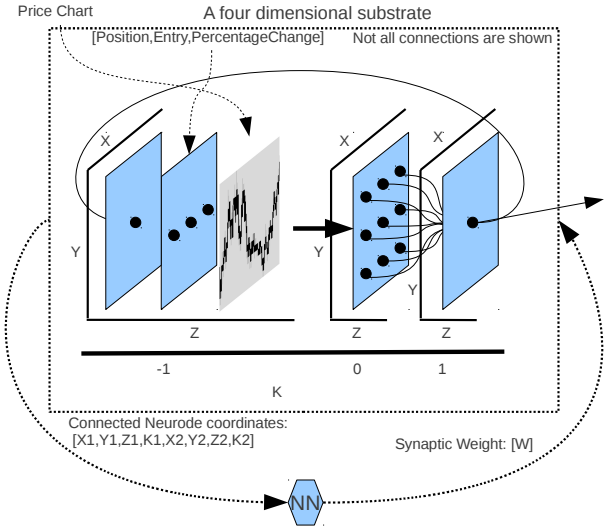

Fig-5.1 A Jordan Recurrent, 4 dimensional substrate encoded NN

To test generalization abilities of the evolved NN based agents, every 500 evaluations the best NN in the population at that time is applied to the 200 data point generalization test. Performing the generalization tests consistently throughout the evolution of the population not only tests the generalization ability of the best NN in the population, but also builds a plot of the general generalization capabilities of that particular encoding and sensor type. This will allow us to get a better idea of whether generalization drops off as the PCI and PLI NNs are over-trained, or whether it improves, or stays the same throughout the training process.

## 6. BENCHMARK RESULTS & DISCUSSION

The general results of the benchmarks are shown in Table-1. The values for this table were computed as follows:
**Training Average Fitness (TrnAvg)** = The average fitness score calculated from the 10 evolutionary runs, reached during the last generation of the population.
**Training Best Fitness (TrnBst)** = The highest achieved fitness amongst the 10 evolutionary runs for that experiment.
**Test Worst Fitness (TstWrst)** = The worst generalization/test fitness achieved amongst the 10 evolutionary runs.
**Test Average (TstAvg)** = The average fitness achieved between all the evolutionary runs during the generalization/test phase.
**Test Standard Deviation (TstStd)** = The standard deviation calculated from all evolutionary runs during the test phase.
**Test Best Fitness (TstBst)** = The best fitness achieved amongst the 10 evolutionary runs during the generalization/test phase.
At the bottom of the table I list the **Buy & Hold** strategy, and the **Maximum Possible** profit results. The Buy & Hold profits are calculated by buying trading the currencies at the very start of the training or testing run respectively, and then trading back at the end. The best possible profit is calculated by looking ahead and trading the currencies only if the profit gained before the trend changes will be greater than the spread.

*Table 1. Benchmark/Experiment Results*

| TrnAvg | TrnBst | Tst Wrst | TstAvg | TstStd | TstBst | Price Vector Sensor Type |
|---|---|---|---|---|---|---|
| 540 | 550 | 225 | 298 | 13 | 356 | [SlidWindow5] |
| 523 | 548 | 245 | 293 | 16 | 331 | [SlidWindow10] |
| 537 | 538 | 235 | 293 | 15 | 353 | [SlidWindow20] |
| 525 | 526 | 266 | 300 | 9 | 353 | [SlidWindow50] |
| 548 | 558 | 284 | 304 | 14 | 367 | [SlidWindow100] |
| 462 | 481 | **214** | 284 | 32 | 346 | [ChartPlane5X10] |
| 454 | 466 | **232** | 297 | 38 | 355 | [ChartPlane5X20] |
| 517 | 527 | **180** | 238 | 32 | 300 | [ChartPlane10X10] |
| 505 | 514 | **180** | 230 | 26 | 292 | [ChartPlane10X20] |
| 546 | 559 | **189** | 254 | 29 | 315 | [ChartPlane20X10] |
| 545 | 557 | **212** | 272 | 36 | 328 | [ChartPlane20X20] |
| 532 | 541 | **235** | 279 | 23 | 323 | [ChartPlane50X10] |
| 558 | 567 | **231** | 270 | 20 | 354 | [ChartPlane50X20] |
| *311* | *N/A* | *N/A* | *300* | *N/A* | *N/A* | ***Buy & Hold*** |
| *N/A* | *704* | *N/A* | *N/A* | *N/A* | *428* | ***Max Possible*** |

First, we note that the generalization results for both, the PCI based NNs and PLI based NNs show profit. The profits are also relatively significant, thus showing that the application of topology and weight evolving artificial neural network systems is viable within this field, and warrants significant further exploration in other time series analysis applications. For example the highest profit reached during generalization, $67 out of the $128 possible when the agent started with 300$, making $100 with 50 leverage based trades, shows that the agent was able to extract 52% of the available profit. This is substantial, but we must keep in mind that though the agents were used on real world data, they were still only trading in a simulated market. It is only after these agents are allowed to trade in real time and using real money, would it be possible to say with certainty that these generalization abilities carry over, and for how many time-steps before the agents require re-training (In the experiment the agents are trained on 800 time steps, and tested on the immediately followed 100 time steps). But looking at the table, the difference between PLI and PCI based NNs do show some strange anomalous.

The PCI based experiment results are particularly surprising. The first thing we notice is that the PCI NN generalization phase's worst performers are significantly worse than those of the PLI based NNs. The PCI based NNs either generalized well during an evolutionary run, or lost significantly. The PLI based NNs mostly kept close to 300 during generalization test phase when not making profit. Also, on *average* the best of PCI are lower than those produced by PLI during generalization. The training fitness scores are comparable for both the PCI and PLI NNs. Another observation we can make is that higher price resolution (X20 Vs. X10) correlates with the PCI NNs to achieving higher profit during generalization testing. And finally, we also see that for both PLI and PCI, generalization achieved by 5 and 100 based windows price windows is highest.

Based on Table-1, at the face of it, it would seem as if the original hypothesis about the effectiveness of PCI NNs, and their expected superior generalization was wrong. But this changes if we now plot the best training fitness vs evaluations, and the best generalization test fitness vs evaluations, as shown in Fig-6.1.

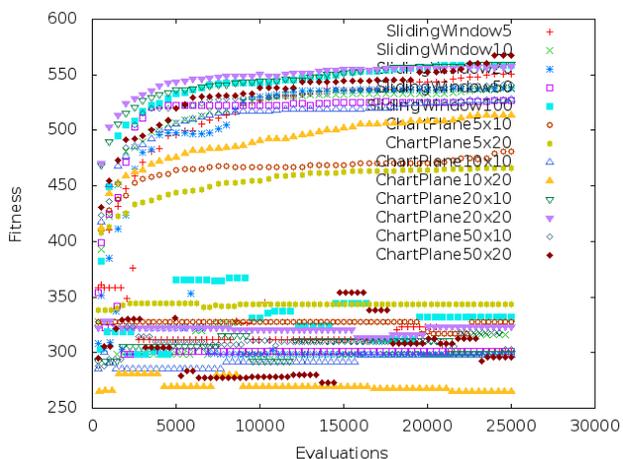

Fig-6.1 PLI & PCI based "Training and Testing Fitness Vs. Evaluations"

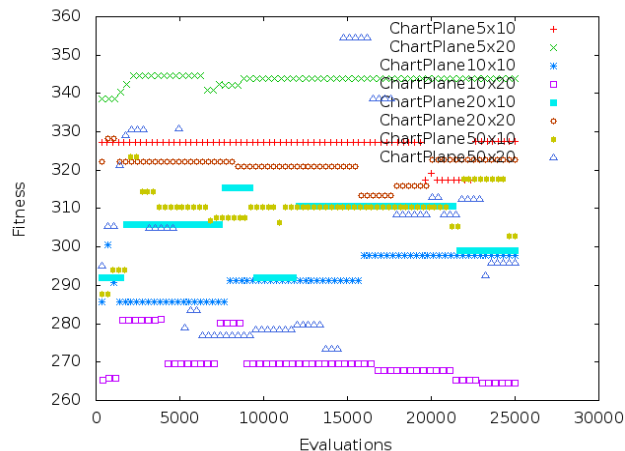

Fig-6.3 PCI based "Generalization Testing Fitness Vs. Evaluations"

Though difficult to see in the above plot, we can make out that though yes the PLI NNs did achieve those generalization fitness scores, they were simply blips during the experiment, occurring a few times, and then disappearing, diving back under 300. On the other hand though, the PCI NNs produced lower profits on average when generalization was tested, **but** they produced those profits consistently, they generalized very well. This is easier to see if we analyze the graph of PLI Generalization Fitness Vs. Evaluations, shown in Fig-6.2, and the PCI Generalization Fitness Vs. Evaluations, shown in Fig-6.3.

Lets now analyze Fig-6.3, the generalization results for just the PCI based NN systems. The story here is very different. Not only there are more consistently higher than 300 generalization fitness scores in this graph, but also they last throughout the entire 25000 evaluations. This means that there are more agents that consistently generalize, reaching the profitability in this graph. Which gives hope that the generalization ability of these PCI NN based systems will carry over to real world trading.

When going through raw data, it was usually the case that for every PLI NN based experiment, only about 1-2 in 10 evolutionary runs had a few agents which generalized for a brief while to scores above 320. On the other hand when going through the PCI NN based experiments, 3-6 out of 10 evolutionary runs had agents generalizing consistently, with scores above 320.

The more conservative PCI NNs are much more consistent. Their generalization stays, and if we look at the above figure, we see that those PCI NNs that have generalization fitness over 300, usually retain it throughout the evaluations. Thus both of the original hypothesis are confirmed. 1. Topology and Weight Evolving Artificial Neural Networks are indeed useful within this field, and their percentage profits are higher than those reported in the references papers which used backprop algorithms in optimization of static topology based NNs. And 2. Geometrical pattern sensitive, price chart input based NNs do work, the NNs learned how to trade currencies based on geometrical patterns within the charts, looking at the trends and patterns, and were able to generalize much better. Thus this new proposed method of evolving geometrical pattern sensitive currency trading agents is viable.

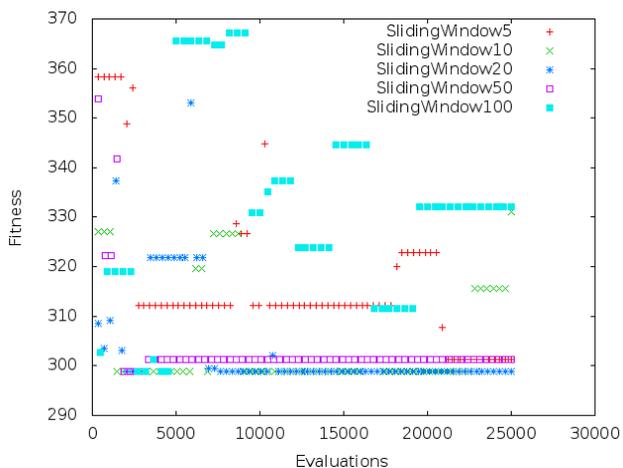

Fig-6.2 PLI based "Generalization Testing Fitness Vs. Evaluations"

If we look at SlidingWindow100, produced by plotting the best generalization scores from the 10 evolutionary runs of that experiment, we see that the score of 367 was achieved briefly, between roughly the evaluation number 5000 and 10000. This means that there was most likely only a single agent out of all the agents, in the 10 evolutionary runs, that achieved this, and then only briefly so. On the other hand, we also see that majority of the points are at 300, which implies that most of the time, the agents did not generalize. And as expected, during the very beginning, evaluations 0 to about 3000, there is a lot more activity amongst all sliding window resolutions, which produce profits. The most stable generalization and thus profitability was shown by SlidingWindow5 and SlidingWindow100, and we know this because in those experiments, there were a lot more fitness scores above 300, consistently. From this, we can extract the fact that during all the experiments, there are only a few agents that generalize well, and do so only briefly, when it comes to PLI based NNs.

But there were a few issues, and when experimenting with PCI NNs, problems did show up. First, the standard substrate hypercube[6] topologies are layer-to-layer feed forward topologies, and fully connected topologies. I did experiment using feedforward substrates, but they could not generalize in this domain, and their produced fitness scores were bellow 290. This might be due to their ability to memorize, but without any recurrent connections, lacking the ability to generalize to previous unseen data, substrate topology based anomalies will need to be analyzed in future work. Also, the fully connected substrate topologies did not generalize with as high a stability as the Jordan Recurrent (JR) topologies used in this paper. The results of using feedforward substrates where each neurode was recurrently connected to itself did not produce results better than a JR substrate either.

I also noted that the low resolution substrates produce at times better results, and do so more often, than their high resolution counterparts... This is due to the fact that by increasing the number of neurodes in one layer, the neurodes in the postsynaptic layer will now have so many inputs that they become saturated, becoming unable to function effectively. Scaling and normalizing the presynaptic vectors for every neurode did not seem to improve this problem significantly, leaving this anomaly to future work as well. These problems make it clear that further experimentation with various substrate topologies is a must.

When analyzing the topology of the best performing PLI NNs, it was clear that they all had one feature in common, they all had a substantial number of recurrent connections. And indeed it was those PLI NNs which used the sliding window vectors of size 5 that seemed to be better at generalizing, which I believe was due to it being difficult for them to evolve simple memorization of signals when using such a small sliding window, which required the evolution of recurrent connections, and which would then help with the ability to generalize. But this is just a hypothesis at the moment.

## 7. Conclusion and Future Work

In this paper I presented the performance, profitability, and generalization of *Price List Input using NNs,* directly encoded, and the *Price Candle-Stick Chart Input using, geometrical pattern sensitive NNs*. I presented a completely new type (to this author's knowledge) of trading and prediction system that uses the actual charts of financial instruments as input, thus letting the evolved NNs take into account the geometrical patterns of the financial data when making predictions and trading. The hypothesis that Topology and Weight Evolving Artificial Neural Network (TWEANN) systems could effectively evolve currency trading agents was shown, based on the generalization results, to be **correct**. Yielding higher profits, and with some agents being able to extract as much as 52% of possible profit during generalization tests. The hypothesis that geometrical pattern sensitive NN systems could indeed trade profitably and generalize much better and more consistently than standard PLI NNs proved to be **correct**. Though at the start it seemed as if the PLI NNs generalized better, after analyzing Fig-6.2 we saw that there were only a few agents, and only for a brief time, which generalized during testing, but they rapidly disappeared. While the PCI NNs generalized consistently during testing, holding profitability throughout 25000 evaluations.

At the same time I noted that it took a lot of experimentation with different types of substrate topologies, and that though Jordan Recurrent topology allowed for the substrate encoded NNs to generalize, others did not fair as well. Thus, more exploration of the various different topologies is needed. Experimentation with free-form substrate topology (where the substrates evolve, with new neurons being integrated and forming new feedforward and recurrent connections over time...) should be undertaken in future work. Furthermore, an expanded set of indicators must also be included in the next phase of testing the use of geometrical analysis based foreign exchange currency trading agents. In this future work, we could have the input hyperplanes be composed of multiple CPIs, having different time-frames, and showing different indicators, and even different currency pairs.

I set out in this paper to show that TWEANN systems have a place in evolving financial instrument trading agents, and the presented experiments proved this to be a correct hypothesis. I also set out to demonstrate a new type of NN based trading agent, one that uses the geometrical patterns within the charts to trade currency pairs, and the demonstration showed that these type of NNs can indeed trade profitably, and generalize better than their standard sliding window input based NN counterparts. The fact that this new approach is effective, gives me hope that it can be applied efficiently to time series analysis in other fields.